%% file: root.tex
\documentclass[letterpaper, 10 pt, conference]{ieeeconf}  

\IEEEoverridecommandlockouts                              

\makeatletter
\def\endthebibliography{%
	\def\@noitemerr{\@latex@warning{Empty `thebibliography' environment}}%
	\endlist
}
\makeatother

\input{def}

\usepackage{graphicx}
\usepackage{epstopdf}
\usepackage{epsfig} 
\usepackage{amsmath} 
\usepackage{amssymb}  
\usepackage{mathtools}
\usepackage{pgfplots}
\usepackage{amsfonts}
\usepackage{url}
\usepackage[]{units}
\usepackage{tikz,everypage}
\usepackage{multirow}
\usepackage{multicol}
\usepackage{lipsum}
\usepackage{subcaption}
\usetikzlibrary{calc}
\usetikzlibrary{positioning}
\graphicspath{{./img/}}
\usetikzlibrary{arrows.meta}
\newcommand{\arr}{-{Latex[length=1mm, width=1mm]}}
\newcommand{\arrd}{{Latex[length=1mm, width=1mm]}-{Latex[length=1mm, width=1mm]}}
\usetikzlibrary{shadings,intersections}
\newcommand{\MINUS}{\scalebox{0.75}[1.0]{\(-\)}}

\title{\LARGE \bf Offset-free Model Predictive Control:\\ A Ball Catching Application with a Spherical Soft Robotic Arm}

\author{Yaohui Huang, Matthias Hofer and Raffaello D'Andrea
\thanks{The authors are members of the Institute for Dynamic Systems and Control, ETH Z\"urich, Switzerland. Email correspondence to Yaohui Huang {\tt\small huangyao@ethz.ch}.}}%
\begin{document}

\maketitle
\thispagestyle{empty}
\pagestyle{empty}
\input{sections/abstract}

\input{sections/introduction}
\input{sections/platform}
\input{sections/modeling}
\input{sections/MPC}
\input{sections/application}
\input{sections/conclusion}
\input{sections/acknowledgment}
\bibliographystyle{IEEEtran}
\bibliography{bibliography}

\end{document}

%% file: def.tex
\newcommand{\idxk}{
\hspace{-0.5pt}(k)\hspace{0.5pt}
}

\newcommand{\idxkmo}{
\hspace{-0.5pt}(k\scalebox{0.75}[1.0]{\(-\)}1)\hspace{0.5pt}
}


%% file: sections/abstract.tex
\begin{abstract}\label{sec:Abstract}
This paper presents an offset-free model predictive controller for fast and accurate control of a spherical soft robotic arm. In this control scheme, a linear model is combined with an online disturbance estimation technique to systematically compensate model deviations. Dynamic effects such as material relaxation resulting from the use of soft materials can be addressed to achieve offset-free tracking. The tracking error can be reduced by 35\% when compared to a standard model predictive controller without a disturbance compensation scheme. The improved tracking performance enables the realization of a ball catching application, where the spherical soft robotic arm can catch a ball thrown by a human.   
\end{abstract}

%% file: sections/introduction.tex
\section{Introduction}\label{sec:Introduction}

Pneumatically actuated soft robots show potential in various applications \cite{polygerinos2017soft}, \cite{gaiser2012compliant}, including prosthetics \cite{pneumaticprosthetichand} and
general automation tasks like pick and place \cite{SoftPickAndPlace}. Their inherent compliance and low
inertia make them safe to work alongside humans \cite{sanan2011physical}, and the use of pneumatic actuation allows for fast maneuvers \cite{mosadegh2014pneumatic}. However, these advantages come at the cost of a more challenging control task. The use of soft materials introduces dynamical effects such as material relaxation that are hard to describe with compact models suitable for control and can lead to tracking offsets and an overall degradation of control performance.\par
Model predictive control (MPC) is an optimization-based control approach: At each time step, the future evolution of the system is predicted based on a model of the system \cite{morari1999model}. The optimal input sequence is computed by solving an optimization problem where a control objective is minimized along the prediction horizon. Thereby, state and input constraints can be explicitly incorporated, making MPC a systematic control approach for handling constraints, while providing an optimal control input for a given objective function.\par
The authors of \cite{gillespie2016simultaneous} apply MPC to a single degree of freedom antagonistic soft inflatable robot. By penalizing the tracking error and the pressure magnitude in the objective function, the resulting model predictive controller can achieve simultaneous position and stiffness control. As a continuation of \cite{gillespie2016simultaneous}, a neural network is used to describe the dynamics of the soft inflatable robot and MPC is applied to the learned model \cite{gillespie2018learning}. To control a six degrees of freedom soft robotic platform, a model predictive controller based on a neural network model is implemented in \cite{hyatt2019model}, where training data is generated from a first principle model.\par
\begin{figure}[!th]
	\centering
	\includegraphics[height=2in]{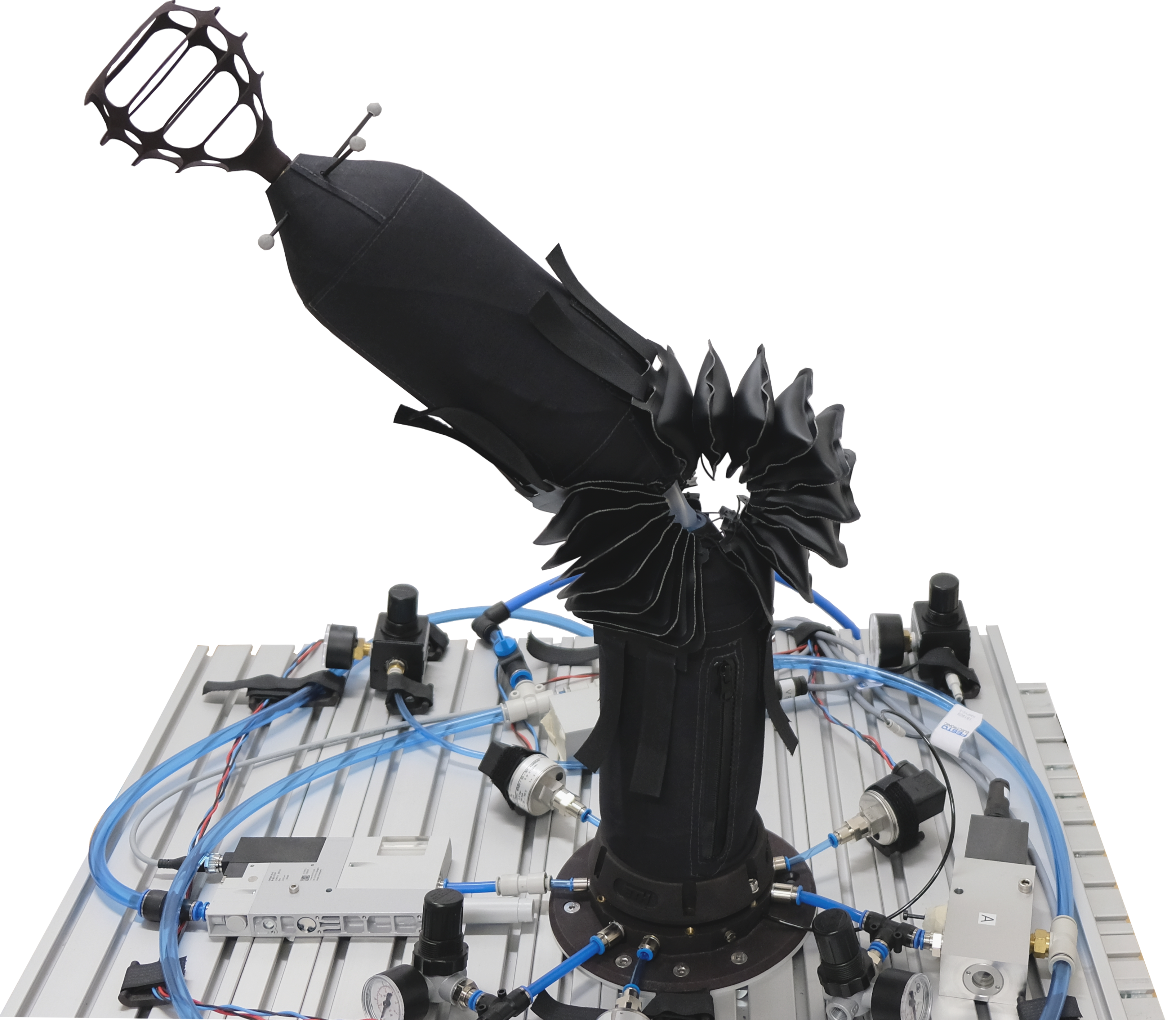}
	\caption{The inflatable soft robotic arm used for the experimental evaluation of the proposed control approach. The system is pneumatically actuated and a 3D printed net is attached to the arm as an end-effector for catching a ball.}
	\label{img:V3photo}
\end{figure}
MPC is capable of rejecting impulse disturbances as it introduces state feedback into the optimization problem in a receding horizon fashion. However, MPC can not track a set point in steady state with zero offset in the presence of persistent model errors or external disturbances. The reason for these tracking offsets is that MPC is not able to correctly predict the system's behaviors without the knowledge of the model errors or the external disturbances. Integral action can be applied to MPC to compensate for occurring offsets as demonstrated in \cite{gillespie2018learning} and \cite{hyatt2019model}. Adaptive control schemes, which adapt model parameters online, can be combined with MPC to eliminate steady state tracking errors, as shown in \cite{hyatt2020model}. The authors combine model reference adaptive control with MPC on a continuum joint robot to compensate for model mismatch. An alternative method to eliminate tracking offsets is the control approach referred to as offset-free MPC (\cite{borrelli2007offset}, \cite{pannocchia2015offset}). Thereby, the system model is augmented with a disturbance state that is estimated online and compensated for in the MPC optimization problem. This allows the elimination of tracking offsets in a systematic way.\par
Ball catching has been extensively studied as a benchmark task to evaluate robotic systems, because it requires fast and accurate control performance. Various examples of ball catching have been demonstrated for rigid body robots (\cite{ExperimentsinRoboticCatching}, \cite{bauml2010kinematically}) and for a robot arm composed of flexible links \cite{link_flexible}. A soft prosthetic hand capable of grasping a ball thrown towards the hand is discussed in \cite{2018elastomeric}. However, to the best of our knowledge, no ball catching application has been demonstrated with a fully soft robotic system that can autonomously catch a thrown ball by moving to the predicted intersection point.\par 
In this work, ball catching is realized with a spherical soft robotic arm (see Fig. \ref{img:V3photo}). As the application requires the controller to track a trajectory that is unknown beforehand, previously explored control approaches such as iterative learning control (as used in \cite{zughaibi2020fast} for realizing a pick and place application) are not feasible for this application. Therefore, an offset-free MPC approach that can deal with an a priori unknown reference trajectory is employed for realizing the ball catching application. \par 
The remainder of this paper is organized as follows: A brief overview of the soft robotic system is given in Section \ref{sec:platform}, followed by the modeling of the robotic manipulator in Section \ref{sec:Modeling}. The offset-free MPC approach is discussed in Section \ref{sec:MPC} and experimentally evaluated through a comparison to a standard MPC approach. The realization of the ball catching application is presented in Section \ref{sec:application} and a conclusion is drawn in Section \ref{sec:conclusion}. 

%% file: sections/platform.tex
\section{Soft Robotic Platform Overview}\label{sec:platform}

\begin{figure}[!t]
	\vspace{5 pt}
	\centering
	\resizebox{0.5\textwidth}{!}{\input{img/angleConvention.tikz}}
	\caption{The left hand plot shows the soft robotic arm and the different components it is composed of. A static link (1) is connected to a movable inflatable link (2) over a soft joint (3). The links are made of lightweight fabric material, and the soft joint is made from silicone rubber. The system is pneumatically actuated by soft bellow actuators (4) that are fabricated from soft material with high tensile strength. A 3D printed net (5) is attached to the movable link for the ball catching application. The middle plot shows the orientation parametrization using the extrinsic Euler angles $\alpha$ and $\beta$. The right hand plot shows the antagonistic configuration of the three symmetrically arranged bellow-type actuators.}
	\label{fig:angleConvention}
\end{figure}
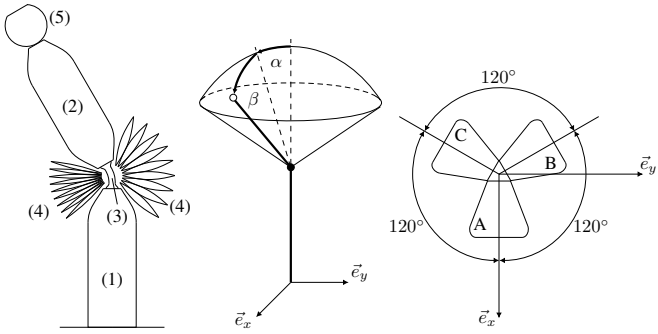

The spherical soft robotic arm originates from \cite{hofer2020design} and an improved version is used in \cite{zughaibi2020fast}. The individual components of the soft manipulator are explained in Fig. \ref{fig:angleConvention} (left). Connected to a flexible soft joint, the movable link can move on a section of a sphere parametrized by the extrinsic Euler angles $\alpha$ and $\beta$ as shown in Fig. \ref{fig:angleConvention} (middle). The three actuators have an antagonistic configuration as shown in Fig. \ref{fig:angleConvention} (right). Adjusting the air pressures of the actuators allows for control of their expansion and consequently the orientation of the movable link.\par 
The angles $\alpha$, $\beta$ are retrieved from an infrared motion capture system running at 200 Hz. The pressure in each actuator is measured with a pressure transducer sampled at 200 Hz and is controlled by a separate proportional–integral–derivative controller (PID controller).  

%% file: img/angleConvention.tikz
\begin{tikzpicture}[scale = 1]
\begin{scope}[scale=1,shift={(2.0859,0.0414)}]


\draw (-1,-0.9) -- (1,-0.9);

\draw[thin] (0,-0.9) -- (0.46,-0.9) -- (0.46,1.1);
\draw[thin] (0.46,1.1) to [out=90,in=300] (0.17,1.75); 
\draw[thin] (0.17,1.75) -- (-0.17,1.75);
\draw[thin] (-0.46,1.1) to [out=90,in=240] (-0.17,1.75); 
\draw[thin] (0,-0.9) -- (-0.46,-0.9) -- (-0.46,1.1);

\begin{scope}[rotate around={210:(0,2)}]
\draw[thin] (-0.46,-0.25) to [out=270,in=120] (-0.17,-0.9); 		
\draw[thin] (-0.17,-0.9) -- (0.17,-0.9) ;	
\draw[thin] (0.17,-0.9) to [out=60,in=270] (0.46,-0.25); 	
\draw[thin] (0.46,-0.25) -- (0.46,1.1);
\draw[thin] (0.46,1.1) to [out=90,in=300] (0.17,1.75); 
\draw[thin] (0.17,1.75) -- (-0.17,1.75);
\draw[thin] (-0.46,1.1) to [out=90,in=240] (-0.17,1.75); 
\draw[thin] (-0.46,-0.25) -- (-0.46,1.1);
\end{scope}

\draw[thin] (-2,5.05) node (v1) {} arc (153.0114:447.0082:0.4);
\draw (-2.0012,5.0453) -- (-1.61,5.2734);

\draw[thin] (0.12,1.75) to [out=110,in=270] (0.06,1.85); 
\draw[thin] (-0.12,1.75) to [out=70,in=270] (-0.06,1.85); 
\draw[thin] (0.06,1.85) to [out=90,in=300] (-0.025,2.165); 
\draw[thin] (-0.06,1.85) to [out=90,in=300] (-0.13,2.105); 
\begin{scope}[rotate around={210:(0,2)}]
\draw[thin] (0.12,1.75) to [out=110,in=270] (0.06,1.85); 
\draw[thin] (-0.12,1.75) to [out=70,in=270] (-0.06,1.85); 
\end{scope}

\foreach \angle in {-42,-28,-14,0,14,28,42,56,70} 
{
	\begin{scope}[rotate around={\angle:(0,2)}]
		\draw[] (0.2,2) to [out=-15,in=195] (1.2,2);
		\draw[] (0.2,2) to [out=15,in=165] (1.2,2);
	\end{scope}	
}

\foreach \angle in {-13,-6,1,8,15,22,29,36,43} 
{
	\begin{scope}[rotate around={\angle:(0,2)}]
\draw[] (-0.2,2) to [out=187,in=-7] (-1.2,2);
\draw[] (-0.2,2) to [out=173,in=7] (-1.2,2);
	\end{scope}	
}

\node at (0,0) {(1)};
\node at (-0.75,3.3) {(2)};
\node at (1.3,1.4) {(4)};
\node at (-1.4,1.3) {(4)};
\draw[thin] (0.1,1.45) -- (0,1.95);
\node at (0.1,1.2) {(3)};
\node at (-1,5) {(5)};

\end{scope}
\begin{scope}[scale=1.1,shift={(5,0)}]

\draw[very thick] (0,0) -- (0,2);
\draw[very thick] (0,2) -- (-0.97,3.16);
\draw[dashed] (0,2.05) -- (0,4.23);
\draw[dashed] (-0.005,2.01) -- (-0.62,4.09);

\draw [\arr, very thick] (0,4.1) arc [radius=1.6, start angle=90, end angle= 112];
\draw [\arr, very thick] (-0.591,3.982) arc [radius=1.25, start angle=132, end angle= 170];
\draw[\arr] (0,0) -- (1,0);
\draw[\arr] (0,0) -- (-0.6,-0.6);
\node at (-0.82,-0.65) {$\vec{e}_x$};
\node at (1.2,0.17) {$\vec{e}_y$};

\node at (-0.25,3.81) {$\alpha$};
\node at (-0.65,3.16) {$\beta$};
\draw [fill] (0,2) circle [radius=0.06];
\draw [] (-1,3.21) circle [radius=0.06];

\coordinate (O) at (0,2);

\draw[thin] (0,2) to [edge label = $$] (-1.5786,3.0748);
\draw[thin] (0,2) -- (1.555,3.0508);

\draw[] (-1.57,3.17) arc [start angle = 140, end angle = 40,
  x radius = 20.4mm, y radius = 26mm];
\draw[densely dashed] (-1.559,3.17) arc [start angle = 170, end angle = 10,
  x radius = 15.8mm, y radius = 3.6mm];
\draw[] (-1.489,3.23) arc [start angle=-200, end angle = 20,
  x radius = 15.8mm, y radius = 3.15mm];

\end{scope}
\begin{scope}[scale=1.1,shift={(4.6153,-0.1188)}]

\draw[\arr] (5.5,2) -- (6.5,2);
\draw[\arr] (4,0.5) -- (4,-0.5);
\draw[thin] (5.5,2) -- (4,2) -- (4,0.5);
\node at (3.8,-0.5) {$\vec{e}_x$};
\node at (6.6,2.15) {$\vec{e}_y$};

\begin{scope}[rotate around={30:(4,2)}]
\draw[thin] (4,2) -- (6,2);
\end{scope}
\begin{scope}[rotate around={150:(4,2)}]
\draw[thin] (4,2) -- (6,2);
\end{scope}
%
\node at (4,3.7) { $120^{\circ}$};
\node at (2.4,1.1) {$120^{\circ}$};
\node at (5.6,1.1) {$120^{\circ}$};
\draw [\arrd,thin]  (4.01,0.5) arc (-90:30:1.5);
\draw [\arrd,thin] (5.309,2.75) arc (29.9992:149.9996:1.5);
\draw [\arrd,thin] (4,0.5) arc (-89.9985:-210.0018:1.5);

\draw (3.5,1.1) -- (3.8,1.88) -- (4.2,1.88) -- (4.5,1.1);
\draw (3.6,0.9) -- (4.4,0.9);
\draw[] (4.5,1.1) to [out=-70,in=0] (4.4,0.9);
\draw[] (3.5,1.1) to [out=-110,in=180] (3.6,0.9);
\node at (3.67,1.13) {A};

\begin{scope}[rotate around={120:(4,2)}]
\draw (3.5,1.1) -- (3.8,1.88) -- (4.2,1.88) -- (4.5,1.1);
\draw (3.6,0.9) -- (4.4,0.9);
\draw[] (4.5,1.1) to [out=-70,in=0] (4.4,0.9);
\draw[] (3.5,1.1) to [out=-110,in=180] (3.6,0.9);
\end{scope}
\node at (4.9,2.2) {B};

\begin{scope}[rotate around={240:(4,2)}]
\draw (3.5,1.1) -- (3.8,1.88) -- (4.2,1.88) -- (4.5,1.1);
\draw (3.6,0.9) -- (4.4,0.9);
\draw[] (4.5,1.1) to [out=-70,in=0] (4.4,0.9);
\draw[] (3.5,1.1) to [out=-110,in=180] (3.6,0.9);
\end{scope}
\node at (3.34,2.69) {C};

\end{scope}
\end{tikzpicture}

%% file: sections/modeling.tex
\section{System Modeling}\label{sec:Modeling}
In this section, the modeling of the soft robotic arm is presented. First, a control allocation strategy that simplifies the modeling process is discussed. Subsequently, a linear state-space model is proposed where its parameters are determined from system identification experiments.  
\subsection{Control Allocation}
The three actuator pressures $(p_A,\, p_B,\, p_C)$ form the inputs to the robotic arm. Beside the two orientation degrees of freedom, there is an additional degree of freedom related to the joint stiffness of the system (see \cite{zughaibi2020fast}, \cite{hofer2020design}). In order to simplify the modeling of the system, an alternative representation of the three inputs is employed using two pressure differences $(\Delta{p}_{\alpha},\, \Delta{p}_{\beta})$ and a lower pressure bound $\bar{p}$ (see \cite{zughaibi2020fast} for a detailed description). The two pressure differences are related to the orientational degrees of freedom and the lower pressure bound to the joint stiffness.\par
The bijective mapping relates the two representations, namely
\begin{equation}
	(\Delta{p}_{\alpha},\, \Delta{p}_{\beta},\, \bar{p}) = \xi(p_A,\, p_B,\, p_C) \text{\,,}
\end{equation}
which is defined by concatenating
\begin{equation}
	\label{}
	\begin{bmatrix}
		\Delta{p}_{\alpha}\\
		\Delta{p}_{\beta}\\
	\end{bmatrix} = T
	\begin{bmatrix}
		\Delta{p}_{AB}\\
		\Delta{p}_{BC}\\
	\end{bmatrix} \quad \text{with} \quad	T = 
	\begin{bmatrix}
		0 & \sqrt{3}/2\\
		\MINUS1 &  \MINUS1/2
	\end{bmatrix}
\end{equation}
and
\begin{equation}
	\label{transformation:absolute to delta}
	\begin{aligned}
		&\Delta{p_{AB}}  = p_{A} \MINUS p_{B} \\
		&\Delta{p_{BC}}  = p_{B} \MINUS p_{C} \text{\,.}
	\end{aligned}
\end{equation}
The lower pressure level is defined as 
\begin{equation}
	\Bar{p}  = \text{min} \{p_{A}, p_{B}, p_{C}\} \text{\,.}
\end{equation}
The mapping $\xi$ roughly aligns the resulting $\Delta{p}_{\alpha}$ and $\Delta{p}_{\beta}$ with the $\alpha$-direction and the $\beta$-direction, respectively. More specifically, for small values of $\alpha$ and $\beta$, a change in $\Delta{p}_{\alpha}$ only causes a deflection in the $\alpha$-direction, and a change in $\Delta{p}_{\beta}$ only causes a deflection in the $\beta$-direction. Therefore, applying the $\xi$ mapping to the actuator pressures and their set points can decouple the system dynamics in the $\alpha$- and $\beta$-directions and simplify the modeling process.\par
The inverse mapping, 
\begin{equation}
 	(p_A,\, p_B,\, p_C) = \xi^{\MINUS1}(\Delta{p}_{\alpha},\, \Delta{p}_{\beta},\, \bar{p}) \text{\,,}
\end{equation}
is given by
\begin{equation}
	\begin{aligned}
    &p_{A} = \text{max}\{\Bar{p},\, \Bar{p}  + \Delta{p_{AB}},\, \Bar{p}  + \Delta{p_{AB}} + \Delta{p_{BC}} \}\\
    &p_{B} = \text{max}\{\Bar{p},\, \Bar{p}  + \Delta{p_{BC}},\, \Bar{p} \MINUS \Delta{p_{AB}} \}\\
    &p_{C} = \text{max}\{\Bar{p},\, \Bar{p}  \MINUS \Delta{p_{BC}},\, \Bar{p}  \MINUS \Delta{p_{AB}} \MINUS \Delta{p_{BC}} \}\\
    &\begin{bmatrix}
    	\Delta{p}_{AB}\\
    	\Delta{p}_{BC}\\
    \end{bmatrix}
    = T^{\MINUS1}
    \begin{bmatrix}
    	\Delta{p}_{\alpha}\\
    	\Delta{p}_{\beta}\\
    \end{bmatrix} \text{\,.}
	\end{aligned}
\end{equation}

\subsection{Inputs and States}
The input vector $u$ of the system is defined as the set points of $\Delta{p}_{\alpha}$ and $\Delta{p}_{\beta}$, namely
\begin{equation}
	\label{eq:input vector}
	u :=
		(\Delta{p}_{\alpha SP},\,
		\Delta{p}_{\beta SP}) \text{\,.}
\end{equation}
The pressure set points applied to the low-level PID controller $p_{ASP}$, $p_{BSP}$, and $p_{CSP}$ are retrieved by applying the inverse mapping $\xi^{\MINUS1}$ to $u$.
The stiffness-related variable $\bar{p}$ is set to a constant value of $\bar{p} = 1.05$ bar throughout this work, as the control of joint stiffness is not relevant in the ball catching application.
\par
The state vector $x$ of the system is defined as
\begin{equation}
	\label{eq:state vector}
	x :=
		(\alpha,\,
		\Dot{\alpha},\,
		\Delta{p}_{\alpha},\,
		\beta,\,
		\Dot{\beta},\,
		\Delta{p}_{\beta})
	 \text{\,.}
\end{equation}
All states in $x$ are computed from sensor data. The angles $\alpha$ and $\beta$ are retrieved from position data provided by the motion capture system, and $\Dot{\alpha}$ and $\Dot{\beta}$ are computed by applying the finite difference method to the angle measurements. The pressure states $\Delta{p}_{\alpha}$ and $\Delta{p}_{\beta}$ are retrieved by applying the $\xi$ mapping to the pressure measurements. 

\subsection{Model Structure}
The proposed model consists of the decoupled arm dynamics describing the motion of the robotic arm and the decoupled pressure dynamics in the $(	\Delta{p}_{\alpha},\,
\Delta{p}_{\beta})$ space. \par  
As a result of the $\xi$ mapping, the dynamics of $\alpha$ and $\beta$ are assumed to be decoupled. The arm dynamics in $\alpha$- and $\beta$-directions can therefore be modeled as two spring-damper systems,
\begin{equation}
	\begin{aligned}
		\label{V3 linear arm model}
		\Ddot{\alpha} &= \MINUS k_\alpha \alpha \MINUS d_\alpha \Dot{\alpha} + h_{\alpha}\Delta{p}_{\alpha}\\
		\Ddot{\beta} &= \MINUS k_\beta \beta \MINUS d_\beta \Dot{\beta} + h_{\beta}\Delta{p}_{\beta} \text{\,,}
	\end{aligned}   
\end{equation}
where $k_\alpha$, $k_\beta$ are the stiffness coefficients, $d_\alpha$, $d_\beta$ the damping coefficients, and $h_{\alpha}$ and $h_{\beta}$ are coefficients that map pressure differences to angular excitations. All coefficients are mass-normalized. Note that the decoupling property of the $\xi$ mapping is only valid for small values of $\alpha$ and $\beta$. Consequently, the decoupled arm model (\ref{V3 linear arm model}) is an approximation of the true dynamics for large angles.\par
Given that the pressure dynamics are controlled in the inner control loops, they are modeled as two decoupled first-order systems. Thereby, interactions with the arm movement are taken into account. In particular, $\Dot{\alpha}$ is assumed to linearly affect $\Delta\Dot{p}_\alpha$, and analogously for $\Dot{\beta}$ and $\Delta\Dot{p}_\beta$. Therefore, the pressure dynamics can be written as,   
\begin{equation}
	\begin{aligned}
		\label{V3 interacting pressure model}
		\Delta\Dot{p}_\alpha &= 1/\tau_{\alpha}(\Delta{p}_{\alpha SP} \MINUS \Delta{p}_{\alpha}) + c_{\alpha} \Dot{\alpha}\\
		\Delta\Dot{p}_\beta &= 1/\tau_{\beta}(\Delta{p}_{\beta SP} \MINUS \Delta{p}_{\beta}) + c_{\beta} \Dot{\beta} \text{\,,}
	\end{aligned}     
\end{equation}
where $\tau_{\alpha}$, $\tau_{\beta}$ are the time constants of the closed-loop pressure dynamics and $c_{\alpha}$, $c_{\beta}$ are related to the interaction with the arm movements.\par

\subsection{System Identification}
 A series of sinusoidal signals with frequencies ranging from 0.5 Hz to 5 Hz are applied to both inputs $\Delta{p}_{\alpha SP}$ and $\Delta{p}_{\beta SP}$ to generate training data to identify the arm dynamics (\ref{V3 linear arm model}). The sinusoidal signals of different frequencies are applied with equal time duration to generate balanced data for each frequency. Step inputs, which excite all frequencies simultaneously, are used to generate training data to identify the pressure dynamics (\ref{V3 interacting pressure model}). The magnitudes of the applied steps cover a large range of the input space $(\Delta{p}_{\alpha SP},\,\Delta{p}_{\beta SP})$. \par 
The recorded data sets contain time series of variables $\alpha$, $\beta$, $\Delta{p}_{\alpha}$, $\Delta{p}_{\beta}$, $\Delta{p}_{\alpha SP}$ and $\Delta{p}_{\beta SP}$. The differentiated values $\Dot\alpha$, $\Ddot{\alpha}$, $\Dot\beta$, $\Ddot{\beta}$, $\Delta\Dot{p}_\alpha$ and $\Delta\Dot{p}_\beta$ are obtained through  differentiation of the spline interpolations of $\alpha$, $\beta$, $\Delta{p}_{\alpha}$ and $\Delta{p}_{\beta}$. All variables are normalized to $[\MINUS1, 1]$. Finally, the parameters in (\ref{V3 linear arm model}) and (\ref{V3 interacting pressure model}) are identified via linear regression.            


%% file: sections/MPC.tex
\section{Offset-free Model Predictive Control}\label{sec:MPC} 
In this section, the implementation of the offset-free MPC approach is discussed. First, the model identified in the previous section is augmented with a disturbance state, and then the disturbance is estimated with a steady state Kalman filter \cite{simon2006optimal}. Second, the tracking target that accounts for the estimated disturbance and ensures offset-free tracking is computed in the target calculation problem. Third, the constraint set for the MPC optimization problem is defined. Finally, the three components are combined to formulate the MPC optimization problem.
The effectiveness of the offset-free MPC approach is evaluated and compared to a standard MPC without a disturbance compensation scheme.        
\subsection{Disturbance Augmentation and Estimation}
The system model combining the arm dynamics (\ref{V3 linear arm model}) and the pressure dynamics (\ref{V3 interacting pressure model}) can be reformulated as
\begin{equation}
	\label{eq: V3 linear Model Reformiulated}
	\begin{aligned}
		\Dot{x} &= 
		\underbrace{\begin{bmatrix}
			0 &\mkern-15mu 1 &\mkern-15mu 0 & & &\\
			\MINUS k_\alpha &\mkern-15mu\MINUS d_\alpha  &\mkern-15mu h_{\alpha} & & &\\
			0 &\mkern-15mu c_{\alpha} &\mkern-15mu \MINUS 1/\tau_{\alpha} & & &\\
			& & &\mkern-15mu 0 &\mkern-15mu 1 &\mkern-15mu 0\\
			& & &\mkern-15mu\MINUS k_\beta &\mkern-15mu\MINUS d_\beta  &\mkern-15muh_{\beta}\\
			& & &\mkern-15mu 0 &\mkern-15mu c_{\beta} &\mkern-15mu \MINUS 1/\tau_{\beta}
		\end{bmatrix}}_{=:A^c} x +
		\underbrace{\begin{bmatrix}
			0 &\\
			0 &\\
			1/\tau_{\alpha} &\\
			&\mkern-15mu 0\\
			&\mkern-15mu 0\\
			&\mkern-15mu 1/\tau_{\beta}\\
		\end{bmatrix}}_{=:B^c}u \text{\,,}
	\end{aligned}
\end{equation}
where $u$ and $x$ are defined in (\ref{eq:input vector}) and (\ref{eq:state vector}).
Assuming that the disturbance acting on the system remains constant across the MPC prediction horizon, the model is augmented with the following constant disturbance $d$,
\begin{equation}
	\label{eq:disturbance augmented model}
	\begin{aligned}
		\Dot{x} &= A^cx + B^cu + d\\
		\Dot{d} &= 0 \text{\,.}
	\end{aligned}
\end{equation}
The augmented model is discretized using the exact discretization method with a sampling time $T_s$, which results in the following discrete time linear model,
\begin{equation}
	\label{discrete dis augmented model}
	\begin{bmatrix}
		x\idxk\\
		d\idxk
	\end{bmatrix} =
	\begin{bmatrix}
		A &E\\
		0   &I
	\end{bmatrix}
	\begin{bmatrix}
		x\idxkmo\\
		d\idxkmo
	\end{bmatrix} +
	\begin{bmatrix}
		B\\
		0
	\end{bmatrix}u\idxkmo \text{\,,}
\end{equation}
where $A$, $B$ and $E$ are the discrete time matrices, $I$ the identity matrix, and $k$ the time index. \par
As discussed in Section \ref{sec:Modeling}, all states in $x$ are retrieved from sensor data, therefore the measurement model of the augmented system is as follows,
\begin{equation}
	\label{measurement model}
	z\idxk = 
	\begin{bmatrix}
		I &0
	\end{bmatrix}	
	\begin{bmatrix}
		x\idxk\\
		d\idxk
	\end{bmatrix}\text{\,,}
\end{equation}
where $z\idxk$ denotes the measurement at time step $k$.\par
We assume additive zero-mean Gaussian noise in the process model (\ref{discrete dis augmented model}) and the measurement model (\ref{measurement model}). The standard steady state Kalman filter formulation \cite{simon2006optimal} is applied to design an estimator for the augmented state, and the resulting estimate is denoted as $(\hat{x}\idxk,\,\hat{d}\idxk)$. The update equations of the steady state Kalman filter is given by,
\begin{equation}
		\begin{bmatrix}
			\hat{x}\idxk\\
			\hat{d}\idxk
		\end{bmatrix}
	= \hat{A}		
	\begin{bmatrix}
		\hat{x}\idxkmo\\
		\hat{d}\idxkmo
	\end{bmatrix} + \hat{B}u\idxkmo + K_{\infty}z(k) \text{\,,}
\end{equation} 
where $K_{\infty}$ is the steady state Kalman filter gain \cite{simon2006optimal} and $\hat{A}$, $\hat{B}$ are defined as follows,
\begin{equation}
	\begin{aligned}
		\hat{A} &= (I \MINUS K_{\infty}\begin{bmatrix}I&0\end{bmatrix})	\begin{bmatrix}
			A &E\\
			0   &I
		\end{bmatrix}\\
	    \hat{B} &= (I \MINUS K_{\infty}\begin{bmatrix}I&0\end{bmatrix})
	    	\begin{bmatrix}
	    	B\\
	    	0
	    \end{bmatrix} \text{\,.}
    \end{aligned}
\end{equation} 
\subsection{Target Calculation}
Given a single desired angle set point $r = (\alpha_{SP}, \,\beta_{SP})$ and the current disturbance estimate $\hat{d}$, the target state and input $(\bar{x},\,\bar{u})$ that account for the estimated disturbance and ensure offset-free tracking  are computed by solving the following linear equations (see \cite{borrelli2007offset} for more details),
\begin{equation}
	\label{eq: V3 linear ss computation}
	\begin{bmatrix}
		A\MINUS{I} &B\\
		H   &0
	\end{bmatrix}
	\begin{bmatrix}
		\bar{x}\\
		\bar{u}
	\end{bmatrix} =
	\begin{bmatrix}
		\MINUS E \hat{d} \,\\
		r
	\end{bmatrix}  \text{\,,}
\end{equation}
where $H$ selects the first and the fourth entries of $\bar{x}$, 
\begin{equation}
	H = \begin{bmatrix}
		1 &0 &0 &0 &0 &0\\
		0 &0 &0 &1 &0 &0
	\end{bmatrix} \text{\,.}
\end{equation}
If no unique solution exists, the pseudo-inverse is applied to compute $(\bar{x},\,\bar{u})$. The resulting target $\bar{x}$ and $\bar{u}$ take into account the disturbance acting on the state evolution and the tracking of the set point.\par
In most tracking tasks, however, the goal is not to track a single set point, but rather a trajectory of desired set points. In this case, the target calculation for a single set point is extended to address a trajectory of set points. At each time step, a set point trajectory denoted as $(r_0, \dots ,r_N)$ is required to be tracked by the MPC controller, where $N$ is the MPC prediction horizon length, and $r_i$ represents the desired set point $i$ time steps ahead of the current time step. A trajectory of target states and inputs is computed by repeatedly solving (\ref{eq: V3 linear ss computation}) for each set point,
\begin{equation}
	\begin{aligned}
	\label{eq: V3 linear ss computation trj}
	\begin{bmatrix}
		A\MINUS{I} &B\\
		H   &0
	\end{bmatrix}
	\begin{bmatrix}
		\bar{x}_{i}\\
		\bar{u}_{i}
	\end{bmatrix} =
	\begin{bmatrix}
		\MINUS E \hat{d} \,\\
		r_i
	\end{bmatrix}\text{\,,} \quad
    i = 0, \dots, N \text{\,,}
	\end{aligned}
\end{equation}
where $\hat{d}$ is assumed to be constant over the prediction horizon, and $(\bar{x}_{i},\bar{u}_{i})$ denotes the target at prediction step $i$. The target trajectory $(\bar{x}_{0},\bar{u}_{0},\dots,\bar{x}_{N},\bar{u}_{N})$ allows the MPC to optimally plan for the future set points and improve trajectory tracking performance.       

\subsection{Constraint Formulation}
\begin{figure}[!t]
	\vspace{5 pt}
	\centering
	\includegraphics{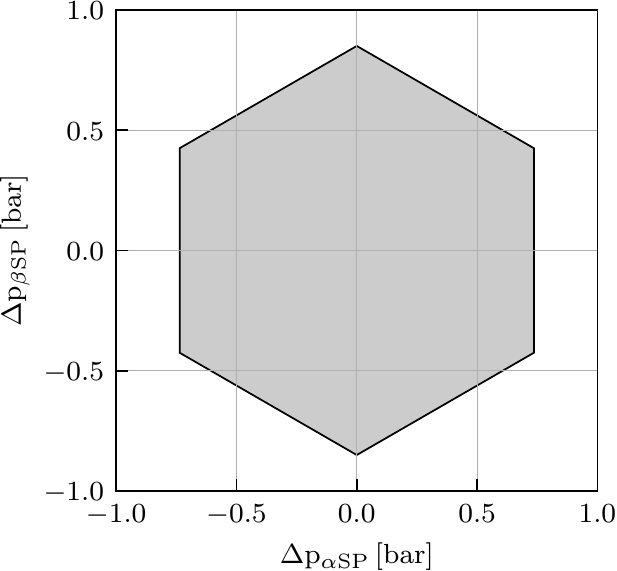}
	\caption{A visualization of the input constraint set $\mathbb{U}$. The feasible inputs are indicated by the gray area.}
	\label{img:V3_Input_Constraint_Set}
\end{figure}

The following constraints apply to the pressure set point of each individual actuator,
\begin{equation}
	\label{eq:V3 Input Constraints}
	\begin{aligned}
		p_{\text{min}} \leq \, &p_{ASP} \leq p_{\text{max}}\\
		p_{\text{min}} \leq \, &p_{BSP} \leq p_{\text{max}}\\
		p_{\text{min}} \leq \, &p_{CSP} \leq p_{\text{max}} \text{\,,}
	\end{aligned}
\end{equation}
where $p_{\text{min}}$ is set to ambient pressure (1.0 bar) and $p_{\text{max}}$ to the maximum allowed pressure (1.9 bar). These constraints limit the pressure set points of the actuators to a safe range. Using the mapping $\xi$, these constraints can be converted to constraints applying to $\Delta{p}_{\alpha SP}$ and $\Delta{p}_{\beta SP}$. A visualization of the resulting constraint set $\mathbb{U}$ is given in Fig. \ref{img:V3_Input_Constraint_Set}. Note that the state $x$ is unconstrained, but a limited set of angles $\alpha$ and $\beta$ is reachable as a consequence of the input constraints. 

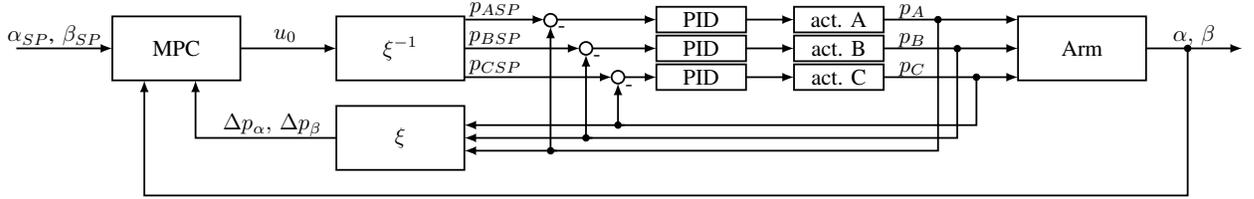
\begin{figure*}[t]
	\vspace{5 pt}
	\centering
	{
		\resizebox{1\textwidth}{!}{
			\input{img/ControlArchitecture.tikz}
		}
		\caption{The cascaded control architecture employed for the robotic arm. The first input from the MPC optimization problem is converted to the pressure set points via the inverse mapping $\xi^{\MINUS1}$. The set points for the actuator pressures are tracked in separate inner control loops at a higher rate.}
		\label{fig:ControlArchitecture}
	}
\end{figure*}   
 
\subsection{MPC Optimization Problem}
Combining the previous building blocks, the MPC optimization problem is formulated as
\begin{equation}
		\label{eq: V3 linear MPC optimization}
		\begin{aligned}
			\min_{x_i,u_i}||x_N\MINUS\bar{x}_{N}||_P&\\
			+ \sum_{i=0}^{N \MINUS 1} ||x_i \MINUS \bar{x}_{i}||_Q
			+ ||u_i \MINUS \bar{u}_{i}||_R &
			+ ||u_i \MINUS u_{i\MINUS1}||_{R_d}\\
			\text{s.t.} \quad x_i = {x}\idxk ,\, u_{i\MINUS1} = u\idxkmo,\,&\text{for $i$ = 0}\\
			x_{i+1} = A x_{i} + B u_i + E \hat{d}\idxk,\,&\text{for $i = 0, \dots ,N\MINUS 1$}\\
			u_i \in \mathbb{U},\, &\text{for $i = 0, \dots ,N\MINUS 1$\,,}
		\end{aligned}
\end{equation}
where $||x||_M$ denotes the induced norm of a vector $x$ by the weighting matrix $M$. 
At time step $k$, ${x}\idxk$ is the measurement of the current state, and
$\hat{d}\idxk$ denotes the disturbance estimate from the steady state Kalman filter. Because the state measurement ${x}\idxk$ has high accuracy and is updated with a higher frequency than the steady state Kalman filter, it has a lower latency and is directly used as the state feedback. The cost matrix $Q$ penalizes a deviation of the state from its target, the matrix $R$ similarly for the input and the cost matrix $P$ a deviation from the terminal target. The last term in the cost function including the cost matrix $R_d$ introduces a cost on the input rate. By penalizing the change between consecutive input values, the smoothness of the resulting input is improved, which limits control action at high frequencies that are beyond the bandwidth of the low-level pressure controllers.\par 
The overall control architecture is shown in Fig. \ref{fig:ControlArchitecture}. The MPC optimization problem (\ref{eq: V3 linear MPC optimization}) is solved in a receding-horizon fashion at each time step. From the first control input $u_0$, the three pressure set points $p_{ASP}$, $p_{BSP}$, and $p_{CSP}$ are retrieved via the $\xi^{\MINUS1}$ mapping. The pressure set points $p_{ASP}$, $p_{BSP}$, and $p_{CSP}$ are then tracked in the inner control loops.
  
\subsection{Trajectory Tracking Experimental Results}
Besides the offset-free MPC discussed above, a standard MPC controller is implemented for comparison. The standard MPC is realized by disabling the disturbance compensation scheme. The prediction horizon length is set to $N = 50$ and the sampling time is set to $T_s\mkern-2mu = \mkern-2mu 0.02s$ for both controllers. The optimization problems are solved using the interior-point-based solver FORCES PRO \cite{FORCESPro}. The controllers are implemented in C++ and are executed on a laptop computer (Intel i7, quad-core, 1.8 GHz).\par
\begin{figure}[t]
	\centering
	\includegraphics{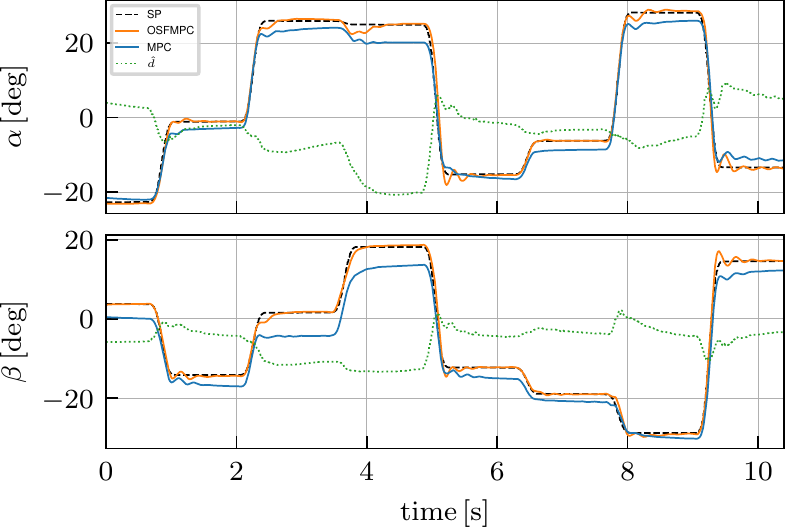}
	\caption{A section of the evaluation trajectory. The black dashed line denotes the set point trajectory (SP), the orange line the trajectory from the offset-free MPC (OSFMPC) and the blue line the trajectory from the standard MPC (MPC). The green dotted line in the upper plot denotes the estimated disturbance in the $\alpha$-direction of (\ref{V3 linear arm model}), and the one in the lower plot denotes the estimated disturbance in the $\beta$-direction. As these two disturbances act in the acceleration directions, they can be interpreted as mass-normalized torques acting on the robotic arm. Without the disturbance compensation scheme, the standard MPC has tracking offsets in the direction of the estimated disturbances. Note that the offsets are not linearly proportional to the disturbances due to the nonlinear relationship between the actuator pressures and the resulting actuation torques. For the tracking of step set points in both the $\alpha$- and $\beta$-directions, the standard MPC shows a maximum steady state offset of $5^{\circ}$ and the offset-free MPC a maximum offset below $0.5^{\circ}$.}
	\label{img:soft step trj}
\end{figure}

 
Both controllers are commanded to track the same angle trajectory consisting of
a ramp trajectory with magnitudes of both angles randomly generated in the interval $[\MINUS30\,\mathrm{deg},\,30\,\mathrm{deg}]$ and angular velocities randomly generated in the interval $[60 \,\mathrm{deg/s},\,300 \, \mathrm{deg/s}]$,
a soft step trajectory with step magnitudes of both angles randomly generated in the interval $[\MINUS30\,\mathrm{deg},\,30\,\mathrm{deg}]$ (a section of this trajectory is shown in Fig. \ref{img:soft step trj}) and
a sinusoidal trajectory with frequencies in the range between $0.5 \,\mathrm{Hz}$ and $3 \,\mathrm{Hz}$. The two controllers are tuned for the best individual tracking performance over the entire trajectory.\par
The Root Mean Square Errors (RMSE) of both $\alpha$- and $\beta$-directions are computed and then averaged. The offset-free MPC controller has an average RMSE of $1.8^{\circ}$, which is 35\% smaller than the average RMSE of $2.8^{\circ}$ from the standard MPC. Moreover, the offset-free MPC can track step set points without offset, while the standard MPC shows static offsets, as shown in Fig. \ref{img:soft step trj}.\par 
We also implemented an MPC with integral action by including the integral of both angles in the state vector. Although this controller shows zero steady state offset, there is a trade off between a fast rise time (high cost on integral states) and little overshoot (small cost on integral states). The overall tracking performance was better than the standard MPC but worse than the offset-free MPC.       

%% file: img/ControlArchitecture.tikz
\usetikzlibrary{arrows}
\begin{tikzpicture}[xscale=1,yscale=1]
\draw[thick]  (-7.5,3.2) rectangle (-6.1,2.8);
\draw[thick]  (-7.5,3.2+0.45) rectangle (-6.1,2.8+0.45);
\draw[thick]  (-7.5,3.2-0.45) rectangle (-6.1,2.8-0.45);
\draw[thick]  (-5.35,3.2) rectangle (-3.95,2.8);
\draw[thick]  (-5.35,3.2+0.45) rectangle (-3.95,2.8+0.45);
\draw[thick]  (-5.35,3.2-0.45) rectangle (-3.95,2.8-0.45);
\draw[thick]  (-12.5,3.5) rectangle (-10.5,2.5) node (v2) {};
\draw[thick]  (-16,3.5) rectangle (-14,2.5);
\draw[thick]  (-1.85,3.5) rectangle (0.15,2.5);
\node at (-15,3) {MPC};
\node at (-11.5,3) {$\xi^{-1}$};
\node at (-6.8,3) {PID};
\node at (-6.8,3.45) {PID};
\node at (-6.8,3-0.45) {PID};
\node at (-4.65,3) {act. B};
\node at (-4.65,3+0.45) {act. A};
\node at (-4.65,3-0.45) {act. C};
\node at (-0.85,3) {Arm};
\draw [-latex, thick](-14,3) -- (-12.5,3);
\draw [-latex, thick](-6.1,3) -- (-5.35,3);
\draw [-latex, thick](-3.95,3) -- (-1.85,3);
\draw [-latex, thick](-3.95,3+0.45) -- (-1.85,3+0.45);
\draw [-latex, thick](-3.95,3-0.45) -- (-1.85,3-0.45);
\draw [-latex, thick](0.15,3) -- (1.65,3);
\draw [-latex, thick](-17.5,3) -- (-16,3);
\draw [-latex, thick] (-12.5,2.1) rectangle (-10.5,1.1);
\node at (-11.5,1.6) {$\xi$};
\draw [-latex, thick](0.8,3) -- (0.8,0.7) -- (-15.5,0.7) -- (-15.5,2.5);
\node at (-8.4306,2.8541) {-};
\node at (-8.978,3.3054) {-};
\node at (-7.9247,2.3889) {-};
\node at (-16.9,3.2) {$\alpha_{SP}$, $\beta_{SP}$};
\node at (-13.3,3.2) {$u_0$};
\node at (-10.0141,3.1577) {$p_{BSP}$};
\node at (-10.0141,3.6077) {$p_{ASP}$};
\node at (-10.0141,2.7077) {$p_{CSP}$};
\node at (0.9,3.2) {$\alpha, \, \beta$};

\draw [-latex, thick](-10.4849,3.45) -- (-9.25,3.45);

\draw [-latex, thick](-9.054,3.45) -- (-7.5,3.45);
\draw [-latex,thick](-10.5,2.55) -- (-8.2,2.55);
\draw [-latex,thick](-8,2.55) -- (-7.5,2.55);
\draw [-latex, thick](-10.5,3) -- (-8.7,3);
\draw [-latex, thick](-8.5,3) -- (-7.5,3);
\draw [-latex, thick](-6.1,3.45) -- (-5.35,3.45);
\draw [-latex, thick](-6.1,2.55) -- (-5.35,2.55);
\draw [-latex, thick](-2.5,2.5574) -- (-2.5,1.8) -- (-8.1,1.8) node (v4) {} -- (-8.1,2.4527);
\draw [-latex, thick](-2.8,3) -- (-2.8,1.6) -- (-8.6,1.6) node (v5) {} -- (-8.6,2.9);

\draw [-latex, thick](-3.1,3.45) -- (-3.1,1.4) -- (-9.154,1.4) -- (-9.154,3.3413);
\node at (-3.5,3.58) {$p_A$};
\node at (-3.5,3.58-0.45) {$p_B$};
\node at (-3.5,3.58-2*0.45) {$p_C$};

\draw [-latex, thick](-8,1.8) -- (-10.5,1.8);
\draw [-latex, thick](-8.5,1.6) -- (-10.5,1.6);
\draw [-latex, thick](-9.054,1.4) -- (-10.5,1.4);
\draw [-latex, thick](-12.5,1.6) -- (-14.7,1.6) -- (-14.7,2.5);
\node at (-13.5,1.8) {$\Delta{p}_{\alpha},\, \Delta{p}_{\beta}$};

\draw [-latex, fill] (-3.101,3.4494) circle (0.05);
\draw [-latex, fill] (-2.799,2.9962) circle (0.05);
\draw [-latex, fill] (-2.5,2.55) circle (0.05);
\draw [-latex, fill] (-8.1,1.8) circle (0.05);
\draw [-latex, fill] (-8.6,1.6) circle (0.05);
\draw [-latex, fill] (-9.1524,1.3988) circle (0.05);
\draw [-latex, fill] (0.8,3) circle (0.05);
\draw [-latex, thick] (-8.6,3)  circle (0.1);
\draw [-latex, thick] (-9.1534,3.4399) circle (0.1);
\draw [-latex, thick] (-8.1,2.55) circle (0.1);

\end{tikzpicture}

%% file: sections/application.tex
\section{Ball Catching Application}\label{sec:application}
In this section, the ball catching application using the offset-free MPC is discussed. First, the intersection point between the thrown ball and the spherical range of the robotic arm is predicted with an estimator. Subsequently, the predicted intersection point is used in a motion planner to generate a trajectory of angle set points for the offset-free MPC. Finally, the set point trajectory guides the controller to catch the ball.      

\subsection{Intersection  Estimator}
An estimator is designed to predict where the thrown ball intersects with the sphere of the arm. The current position of the ball is measured using the motion capture system. An extended Kalman filter \cite{simon2006optimal} is applied to smooth the ball position measurements and estimate the velocity of the ball as well as the aerodynamic drag coefficient. The predicted intersection point between the ball and the spherical range of the arm is computed by forward-predicting the ball model. \par
The ball is modeled as a point mass under the
influence of gravity and aerodynamic drag where spin is neglected \cite{muller2011quadrocopter}. With the ball's position denoted as ${r_B}$, the equation of motion is given by,
\begin{equation}
	\Ddot{{r}}_B = {g} \MINUS K_D ||\Dot{{r}}_B||\Dot{{r}}_B \text{\,,}
\end{equation}
where $K_D$ is the constant aerodynamic drag coefficient, and ${g} = [0,\,0,\,\MINUS9.81]^T$ denotes the gravity vector. As the aerodynamic drag coefficient $K_D$ varies for different balls \cite{muller2011quadrocopter}, it is included in the state to be estimated online. The ball state is defined as
\begin{equation}
	x_B = (r_B,\, \Dot{r}_B,\, K_D) \text{\,.}
\end{equation}
The dynamic model of the ball is formulated as
\begin{equation}
	\label{eq: ball's dynamic model}
	\Dot{x}_B = 
	\begin{bmatrix}
		\Dot{r}_B\\
		{g} \MINUS K_D ||\Dot{r}_B||\Dot{r}_B\\
		0
	\end{bmatrix} \text{\,.}
\end{equation}
The measurement model of the ball is given by
\begin{equation}
	\label{eq: ball's measurement model}
	z_B = [{I}_{3\times3} \quad {0}_{3\times4}]\,x_B \text{\,,}
\end{equation}
where $z_B$ denotes the measurement of the ball position. The dynamic model (\ref{eq: ball's dynamic model}) is discretized using the Runge-Kutta 4th order method with a sampling time of 0.005 s. With the discretized ball model and the measurement model (\ref{eq: ball's measurement model}), the standard extended Kalman filter formulation is applied to design an estimator for the ball state $x_B$.\par
\begin{figure}[t]
	\centering
	\resizebox{0.45\textwidth}{!}{\input{img/MotionPlanning.tikz}}
	\caption{The intersection point $(\tilde{\alpha},\,\tilde{\beta})$ is predicted using the ball's state estimate and its forward prediction. In the motion planner, the shortest path between the current end-effector position $(\alpha,\,\beta)$ and the predicted intersection point $(\tilde{\alpha},\,\tilde{\beta})$ is computed, which lies on the great circle of the sphere. The shortest path is denoted as the dashed line between these two positions.}
	\label{fig:MotionPlanning}
\end{figure}
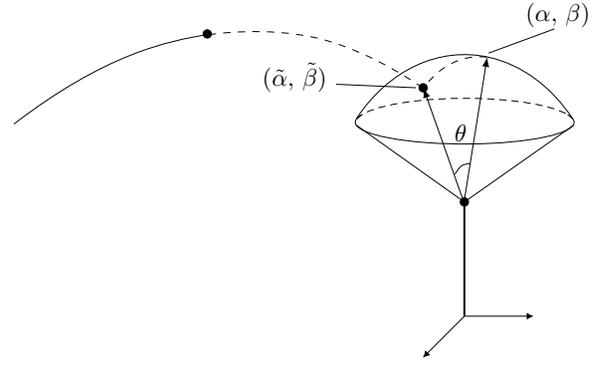

Once the ball is detected in the air, the extended Kalman filter starts to update. Given the estimate of the ball state from the filter, the state is forward predicted until the ball position intercepts with the sphere, as shown in Fig. \ref{fig:MotionPlanning}. The predicted intersection point is expressed in the angular space as $(\tilde{\alpha},\,\tilde{\beta})$.
\subsection{Motion Planner}
At each time step, a desired set point trajectory that connects the current end-effector position $(\alpha,\,\beta)$ and the estimated intersection point $(\tilde{\alpha},\,\tilde{\beta})$ is commanded to the MPC controller. Although it is possible to command the offset-free MPC to directly track the estimated intersection point $(\tilde{\alpha},\,\tilde{\beta})$ as a single step, planning a smoother trajectory for the controller can reduce overshoot and consequently increase the success rate of catching a ball. As the end-effector is moving on a sphere, the shortest path between $(\alpha,\,\beta)$ and $(\tilde{\alpha},\,\tilde{\beta})$ is the circular arc that passes these two positions and lies on a great circle \footnote{The great circle of a sphere is the intersection of the sphere and a plane that passes through the center point of the sphere.} of the sphere (see Fig. \ref{fig:MotionPlanning}). The angle of this circular arc is denoted as $\theta$. By setting a constant angular velocity $\omega_{SP}$, this circular arc is divided into $M$ pieces with equal length, where $M = \theta/(\omega_{SP}T_s)$. If $M$ is not an integer, ceiling is applied. The evenly divided positions on the arc are denoted as $(r_0, \dots, r_M)$,
where $r_0 = (\alpha,\,\beta)$ and $r_M = (\tilde{\alpha},\,\tilde{\beta})$. It is worth emphasizing that the trajectory $(r_0, \dots, r_M)$ is the shortest path between the current arm position and the predicted intersection point. Note that this path does not coincide with a straight line in the $\alpha$-$\beta$-plane.
The first $N+1$ elements of $(r_0, \dots, r_M)$ are set as the set point trajectory for the offset-free MPC. If M is smaller than $N$, the last set point $r_M$ is repeated. In practice, $\omega_{SP}$ is set to $240 \,\mathrm{deg/s}$ to generate a fast trajectory for ball catching. The angular velocity of $240 \,\mathrm{deg/s}$ has been assessed in the trajectory tracking experiment shown in Section \ref{sec:MPC}, and it is  within the tracking capability of the offset-free MPC.
\begin{figure}[t]
	\vspace{5 pt}
	\centering
	\begin{subfigure}{.245\textwidth}
		\centering
		\includegraphics{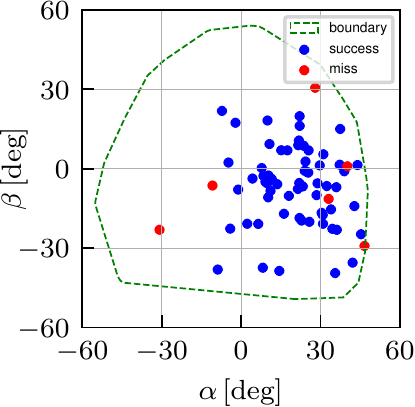}
	\end{subfigure}%
	\begin{subfigure}{.245\textwidth}
		\centering
		\includegraphics{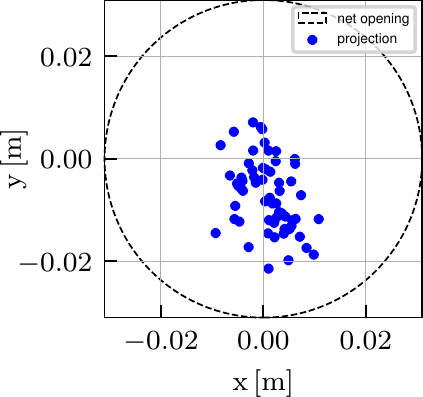}
	\end{subfigure}
	\caption{Statistical analysis of the ball catching application. In the left hand plot, the intersection points are expressed in the angular space. The green dashed boundary denotes the reachable range of the robotic arm given the input constraints listed in Section \ref{sec:MPC}, which is identified from experiments. The blue points are the intersection points of the successful catches, while the red points are the intersection points of the unsuccessful ones. Because the balls are thrown from a fixed position, the intersection points concentrate on a corner of the angular space. In the right hand plot, the intersection points of all the successful throws are projected on to the plane of the net opening.}
	\label{img:statistic analysis}
\end{figure}

\begin{figure}[t]
	\vspace{7 pt}
	\centering
	\includegraphics{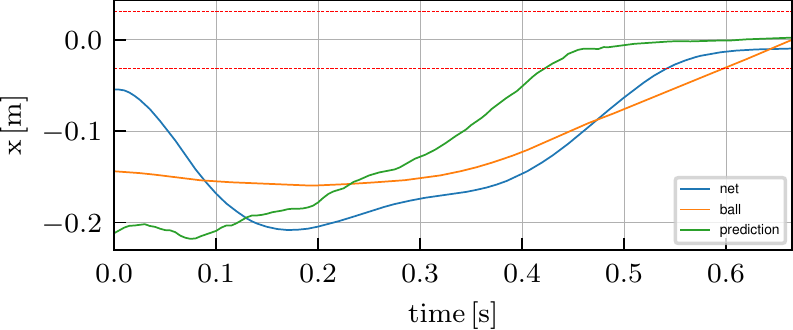}
	\caption{A successful catch when a wind gust disturbance acts on the ball. The $x$ coordinate is expressed in the world frame, with the origin being re-centered to the true intersection point. The blue line denotes the position of the net opening, the orange line the position of the thrown ball, and the green line the predicted intersection point between the ball and the sphere. The red dashed lines denote $\pm31$ mm (the radius of the net opening) from the true intersection point. As the predicted intersection point is used to generate the set point trajectory for the offset-free MPC, the net is controlled toward the predicted intersection point and eventually catches the ball.}
	\label{img:BallCatch_MotionPlanner_Windy}
\end{figure}
\subsection{Experimental Results}
The same controller settings are used for the ball catching experiments as for the evaluation experiment shown in Fig. \ref{img:soft step trj}.
For a statistical analysis of the success rate and the accuracy, the ball catching experiments where a person throws the ball at a fixed position two meters away from the robotic arm are repeated multiple times. The throws that do not intercept with the sphere are not considered in the analysis because they are physically impossible for the robotic arm to catch. The experimental results are shown in Fig. \ref{img:statistic analysis} (left). As the ball is thrown from the positive to the negative $\alpha$-direction, a throw may align with the net opening poorly with a trajectory nearly tangential to the sphere if it lands in the region with negative $\alpha$.
Additionally, a throw may collide with the net's frame when it lands closer to the boundary of the system's range of motion. As a result, these regions have a lower success rate. Among 68 throws, 62 throws were successfully caught by the robotic arm while six throws were not, which equals a success rate of 91\%.
In Fig. \ref{img:statistic analysis} (right), the intersection points of the successful throws are projected on to the plane of the net opening. The average distance from these projected points to the center of the net opening is 9.7 mm, around one-third of the radius of the net opening (31 mm).\par  
In order to test the robustness against disturbances of the proposed control approach, ventilation fans are used to generate wind gusts acting as disturbances in the ball catching experiment. As the predicted intersection point and the set point trajectory from the motion planner are recomputed at every time step, it allows the controller to reject disturbances applied on the flying ball. Because the wind mainly acts in the positive $x$-direction, the predicted intersection point is constantly re-adjusted in this direction, as shown in Fig. \ref{img:BallCatch_MotionPlanner_Windy}. When the ball gets closer to the sphere, the predicted intersection point becomes more accurate. As a result, the net is controlled toward the true intersection position and eventually successfully catches the ball. A video showing the ball catching experiment under the wind gust disturbance is available in the video attachment (https://youtu.be/b8ov1Jzd89k).\par  
We also tried to implement the ball catching application with the standard MPC. However, the success rate is below 30\% for the same experiment as shown in Fig. \ref{img:statistic analysis} due to the significant tracking offsets.                

%% file: img/MotionPlanning.tikz

\usetikzlibrary{arrows}
\begin{tikzpicture}
	\draw[thick] (-1.35,0.15) -- (-1.35,1.8);
	
	\draw[\arr] (-1.35,0.15) -- (-0.35,0.15);
	\draw[\arr] (-1.35,0.15) -- (-1.95,-0.45);
	
	\draw [fill] (-1.35,1.8) circle [radius=0.06];
	
	\coordinate (O) at (-1.35,1.8) {};
	
	\draw[thin] (-1.35,1.8) to [edge label = $$] (-2.911,2.91);
	\draw[thin] (-1.35,1.8) node (v1) {} -- (0.229,2.91);
	
	\draw[] (-2.911,3) arc [start angle = 140, end angle = 40,
	x radius = 20.4mm, y radius = 26mm];
	\draw[densely dashed] (-2.9,3) arc [start angle = 170, end angle = 10,
	x radius = 15.8mm, y radius = 3.6mm];
	\draw[] (-2.83,3.06) arc [start angle=-200, end angle = 20,
	x radius = 15.8mm, y radius = 3.15mm];
	
	\draw  plot[smooth, tension=.9] coordinates {  (-5.0513,4.2257)};
	\draw[dashed]  plot[smooth, tension=.9] coordinates { (-5.041,4.2162) (-3.4299,4.1561) (-1.941,3.44)};
	\draw [fill] (-5.0615,4.2271) circle [radius=0.06];

\draw [-latex](-1.3545,1.802) -- (-1.0207,3.8929) node (v2) {};

\draw [dashed](-1.0284,3.8929) .. controls (-1.6317,3.9547) and (-1.941,3.4366) .. (-1.941,3.4366);
\draw [fill] (-1.9455,3.4495) circle [radius=0.06];
\draw [latex-](-1.941,3.4443) -- (-1.3622,1.8098);

\draw (-5.0383,4.2206) .. controls (-5.6513,4.1257) and (-6.5513,3.9257) .. (-7.8513,2.9257);
\node (v3) at (-3.8,3.6) {$(\tilde{\alpha},\,\tilde{\beta})$};
\node at (0,4.5) {$ (\alpha,\,\beta)$};

\draw  plot[smooth, tension=.7] coordinates {(-1.2629,2.3577) (-1.3932,2.3205) (-1.4862,2.1995)};
\node at (-1.4,2.8) {$\theta$};
\draw (-2.05,3.45) -- (-3.2,3.5);
\draw (-1,3.95) -- (-0.05,4.3);
\end{tikzpicture}

%% file: sections/conclusion.tex
\section{Conclusion}\label{sec:conclusion}
An offset-free MPC controller is implemented on a soft robotic arm. This controller employs a simple linear model and compensates for the residual model errors with a disturbance estimation scheme. In the trajectory tracking evaluation, it outperforms the standard MPC with a 35\% smaller tracking error. Furthermore, the offset-free MPC is used to realize a ball catching application. It is shown that this controller can successfully catch a ball while rejecting external disturbances, which demonstrates the responsiveness, accuracy, and robustness of the control approach proposed. These results indicate that the offset-free MPC can cope with control challenges associated with the use of soft materials despite the simplicity of the employed model.\par
Currently, the stiffness-related variable $\bar{p}$ is set to a constant value in modeling and control, which limits the joint stiffness of the soft robotic arm. Future work will apply offset-free MPC to a model parametrized by $\bar{p}$ for stiffness control of the robotic manipulator.          

%% file: sections/acknowledgment.tex
\section*{Acknowledgment}
The authors would like to thank Jasan Zughaibi, Michael Egli, Matthias Müller, and Helen Hanimann for their contributions to this work.  